\documentclass[10pt,twocolumn,letterpaper]{article}

\usepackage[pagenumbers]{cvpr} 

\usepackage[accsupp]{axessibility}  
\usepackage{graphicx}
\usepackage{amsmath}
\usepackage{amssymb}
\usepackage{booktabs}

\usepackage{multirow}
\usepackage{arydshln}
\usepackage{nicefrac}
\usepackage{xspace}
\usepackage[dvipsnames]{xcolor}

\def\eg{\emph{e.g}\onedot} \def\ie{\emph{i.e}\onedot} 
\def\etal{\emph{et al.}}

\newcommand{\ER}{\emph{ER}\xspace}
\newcommand{\ERMinusMinus}{\emph{ER-{}-}\xspace}
\newcommand{\ERPlusPlus}{\emph{ER++}\xspace}

\usepackage[pagebackref,breaklinks,colorlinks]{hyperref}

\usepackage[capitalize]{cleveref}
\crefname{section}{Sec.}{Secs.}
\Crefname{section}{Section}{Sections}
\Crefname{table}{Table}{Tables}
\crefname{table}{Tab.}{Tabs.}

\def\cvprPaperID{5111} 
\def\confName{CVPR}
\def\confYear{2022}

\newcommand\blfootnote[1]{%
  \begingroup
  \renewcommand\thefootnote{}\footnote{#1}%
  \addtocounter{footnote}{-1}%
  \endgroup
  }
\begin{document}

\title{Real-Time Evaluation in Online Continual Learning:\\A New Hope}

\author{Yasir Ghunaim$^{1*}$
\and
Adel Bibi$^{2*}$
\and
Kumail Alhamoud$^1$
\and
Motasem Alfarra$^1$
\and
Hasan Abed Al Kader Hammoud$^1$
\and
Ameya Prabhu$^2$
\and
Philip H.S. Torr$^2$
\and
Bernard Ghanem$^1$\vspace{0.1cm}
\and 
$^1$ King Abdullah University of Science and Technology (KAUST) $^2$ University of Oxford
}
\maketitle

\begin{abstract}
Current evaluations of Continual Learning (CL) methods typically assume that there is no constraint on training time and computation. This is an unrealistic assumption for any real-world setting, which motivates us to propose: a practical real-time evaluation of continual learning, in which the stream does not wait for the model to complete training before revealing the next data for predictions. To do this, we evaluate current CL methods with respect to their computational costs. We conduct extensive experiments on CLOC, a large-scale dataset containing 39 million time-stamped images with geolocation labels. We show that a simple baseline outperforms state-of-the-art CL methods under this evaluation, questioning the applicability of existing methods in realistic settings. In addition, we explore various CL components commonly used in the literature, including memory sampling strategies and regularization approaches. We find that all considered methods fail to be competitive against our simple baseline. This surprisingly suggests that the majority of existing CL literature is tailored to a specific class of streams that is not practical. We hope that the evaluation we provide will be the first step towards a paradigm shift to consider the computational cost in the development of online continual learning methods.\blfootnote{* Equal Contribution.}\blfootnote{Correspondence to: \texttt{yasir.ghunaim@kaust.edu.sa}}
\blfootnote{Code: \href{https://github.com/Yasir-Ghunaim/RealtimeOCL}{github.com/Yasir-Ghunaim/RealtimeOCL}}

    
\end{abstract}

\section{Introduction}
\label{sec:intro}

\begin{figure*}
\includegraphics[width=\linewidth]{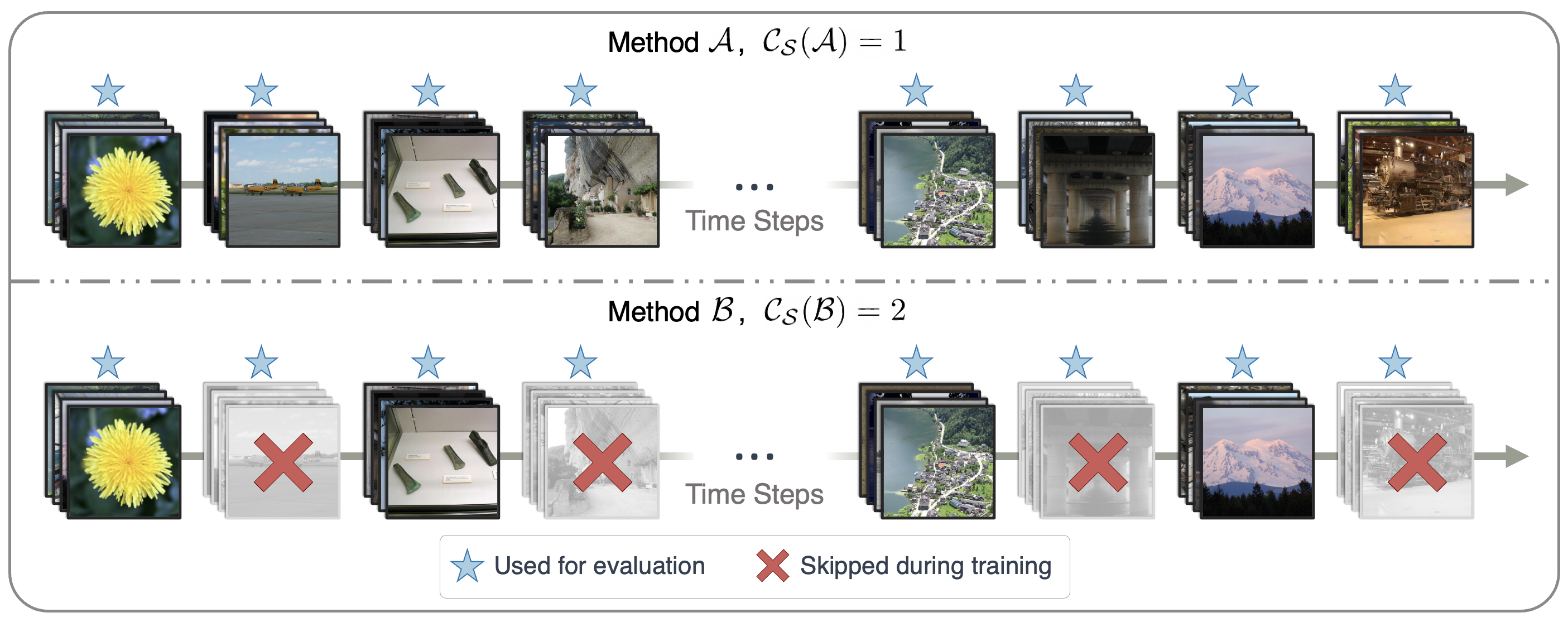}
\vspace{-0.6cm}
\caption{\textbf{OCL Real-Time Evaluation Example.} We show an example of real-time evaluation, using the CLOC dataset~\cite{cai2021online}, of two different OCL methods $\mathcal{A}$ and $\mathcal{B}$. Method $\mathcal{B}$ is twice as slow as method $\mathcal{A}$. Both methods are evaluated on every incoming sample. Since $\mathcal{A}$ has a stream-model relative complexity of one, \ie $\mathcal{C}_\mathcal{S}(\mathcal{A}) = 1$, it is able to train on all the stream samples. In contrast, $\mathcal{B}$, which has a relative complexity of two, requires two time steps to train on a single stream batch. Thus, $\mathcal{B}$ only trains on half of the stream samples.}\vspace{-0.3cm}
\label{fig:pull_figure}
\end{figure*}

Deep Neural Networks~(DNNs) have demonstrated impressive success in solving complex tasks \cite{he2016resnet, liang2015rcnn, schmidhuber2015deep} when trained offline, for several passes, over large well-curated labeled datasets. However, in many real-world scenarios, data is only available in the form of a stream with a changing distribution. Due to this challenge, there has been a growing interest in the problem of learning from a time-varying stream, also known as Continual Learning~(CL), which is a key challenge for DNNs due to a phenomenon known as \emph{catastrophic forgetting}\cite{mccloskey1989catastrophic, french1999catastrophic}. In particular, when a DNN is trained with data from a new distribution, the DNN performance significantly drops on previously learned data.

While mitigation efforts have been proposed, \eg through regularizing the training~\cite{kirkpatrick2017overcoming, zenke2017continual, aljundi2018memory}, replaying previously seen examples~\cite{rebuffi2017icarl, isele2018selective, chaudhry2019continual}, and many other approaches~\cite{rusu2016progressive, fernando2017pathnet, wang2022learning}, current evaluation approaches are still far from real-world scenarios. For example, the majority of literature is on \emph{Offline Continual Learning}, under which methods are allowed unlimited budget, both time and computation. Furthermore, the majority of CL evaluations are conducted on \emph{small-scale} datasets with well-defined temporal distribution boundaries in the form of learning a sequence of tasks.

To that end, there has recently been a growing interest in the more realistic setting -- \emph{Online Continual Learning} (OCL). In such a setup \cite{lopez2017gradient, aljundi2019gradient, hu2021one, alfarra2022simcs}, CL methods are restricted to a single training pass over a shuffled split of existing offline CL benchmarks. This is certainly a step forward towards resolving some of the unrealistic assumptions of offline CL. However, current evaluations do not sufficiently address the challenges of real-time learning for high-throughput streams with rapid distribution changes.

To illustrate this, consider the problem of continuously learning a Twitter stream where 350K tweets are uploaded per minute on various trending topics \cite{sayce_2022}. Every uploaded tweet needs to be predicted with a DNN for misinformation and hate speech, among other things, while simultaneously learning and adapting to them. Given the scale at which data is being updated, there is an inherent key limitation on the time and computational budget affordable to learning incoming tweets, an aspect that is often overlooked in the prior art from the OCL literature.
Consider an OCL method that is 10 times slower than the Twitter high throughput stream, \ie, it takes 10 minutes to train on one minute worth of tweets (350K tweets). This inefficiency results in an accumulation of $\sim$3.1 million new samples that need to be predicted and trained on. Since it is not acceptable to pause all tweets from appearing online until the method training is complete, predictions for all new samples will be performed with an older version of the model. This poses a key challenge where efficient learning from streams becomes necessary. This is because slow-training OCL methods can result in subpar performance, as they resort to predicting new stream data using an older model. This behavior worsens for streams that experience a faster change in distribution.

In this paper, we propose a real-time evaluation protocol for OCL that factors in training computational complexity. Given a stream, consider an OCL method $\mathcal{A}$ that is as fast as the stream, \ie, $\mathcal{A}$ can train on every step of revealed data before the stream presents new samples. Then, if an OCL $\mathcal{B}$ is twice as expensive as $\mathcal{A}$, then $\mathcal{B}$ will update the model for evaluation every other stream step, \ie, the model will be updated half the number of times compared to $\mathcal{A}$. Figure \ref{fig:pull_figure} illustrates our proposed real-time evaluation. This is in contrast to all prior art \cite{aljundi2019gradient, aljundi2019online, caccia2022new}  that (\textbf{1}) unreasonably allows an unlimited computational budget to train on any given stream data, and (\textbf{2}) unfairly compares OCL methods despite having different training complexity levels. Using our real-time evaluation protocol, we benchmark many existing OCL methods against a simple and inexpensive baseline, which mitigates forgetting by simply storing and replaying recently seen samples. 

\noindent\textbf{Contributions.} We summarize our conclusions as follows: \textbf{(1)} We show that under our practical real-time evaluation, our simple baseline outperforms \emph{all} the considered methods from the OCL literature, including recent SOTA approaches like ACE~\cite{caccia2022new}. \textbf{(2)} We consider a complementary setup where the stream is as slow as the most training-expensive OCL method and compare that method against the compute-equivalent baseline. Under this computationally normalized setting, we find that the compute-equivalent baseline outperforms all existing methods. \textbf{(3)} Our experiments are consistent, holding for all the considered continual learning strategies, and extensive, amounting to more than 2 GPU-months. Our results highlight that the current progress in OCL needs to be rethought and a paradigm shift is needed.
We hope our work will lead to a new direction for continual learning that takes into account the computational cost of each method.

\section{Related Work}

We briefly describe the current literature on offline and online continual learning. For a more comprehensive overview of the literature, we refer the reader to the detailed surveys by De Lange \etal \cite{de2021continual_survey} and Mai \etal \cite{mai2022online}.

\vspace{2pt}\noindent\textbf{Offline Continual Learning.} Traditional continual learning strategies, which aim to mitigate forgetting, can be organized into three families of methods. \emph{(i) Replay-based} methods store input samples, or alternatively learn to generate representative samples, while the model learns from the data stream. Later, the method retrains the model on the replay samples while the model learns from new data~\cite{lopez2017gradient, shin2017continual, maracani2021recall, villa2022vclimb, villa2022pivot}. \emph{(ii) Regularization} methods avoid the cost of storing samples and simply modify the model loss objective to regularize the training. While some of these methods penalize changes to important model parameters~\cite{kirkpatrick2017overcoming, aljundi2018memory, chaudhry2018riemannian}, other methods regularize the training by distilling knowledge from a model trained on past stream samples~\cite{li2017learning, smith2021always, gao2022rdfcil}. \emph{(iii) Parameter isolation} methods train specific parameters for each task while freezing the parameters that are related to other tasks~\cite{rusu2016progressive, mallya2018packnet, hu2019overcoming}. Despite the progress made by these methods, they assume an offline stream that allows many passes over each continual learning task. Concurrent work Prabhu \etal \cite{prabhu2023computationally} alleviates this problem by imposing computational budget constraints and finds that simple methods based on experience replay outperform most prior continual learning works. In contrast, we study the more pragmatic setup, where the stream reveals data in real time.

\vspace{2pt}\noindent\textbf{Online Learning for Reduced Forgetting.}
OCL was defined with a protocol where training data is only seen once in a sequence of labeled tasks~\cite{lopez2017gradient}.
To reduce catastrophic forgetting, the field initially progressed towards better gradient-based constraints like GEM~\cite{lopez2017gradient} and AGEM~\cite{chaudhry2018efficient}. RWalk\cite{chaudhry2018riemannian} quantified forgetting and provided a more efficient approach to limit changes in important model parameters.  TinyER \cite{chaudhry2019continual} rediscovered the effectiveness of experience replay, and HAL \cite{chaudhry2020using} enhanced it with learned anchors. However, this setup assumes the availability of an oracle during test time to determine which classifier head to use for inference~\cite{van2019three, mai2022online}. Additionally, the benchmarks in this setup often have large constraints on the replay buffer size~\cite{chaudhry2018efficient, chaudhry2019continual}. Due to these limitations, the class-incremental scenario started to gain attention, which is a more realistic setting for OCL.

In class-incremental online continual learning, benchmarks \cite{aljundi2019online, lee2020neural} relaxed the need for task descriptors at test time, and introduced significantly larger buffer sizes in comparison to \cite{chaudhry2018efficient}. Multiple directions have emerged to tackle catastrophic forgetting in this setup. They can be classified into few groups. \emph{(i) Regularization-based} approaches modify the classification objective to preserve previously learned representations or encourage more meaningful representations  \eg DER \cite{buzzega2020dark}, ACE \cite{caccia2022new}, and CoPE \cite{de2021continual}. \emph{(ii) Sampling-based} techniques focus on the optimal selection and storing of the most representative replay memory during online training, \eg  GSS \cite{aljundi2019gradient}, OCS \cite{yoon2022online}, CBRS \cite{chrysakis2020online}, CLIB \cite{koh2021online}, and InfoRS \cite{sun2022information}. Alternatively, some sampling methods focus on better memory retrieval strategies that reduce forgetting, \eg MIR \cite{aljundi2019online}, ASER \cite{shim2021online}, and GMED \cite{jin2020gradient}. (iii) In other approaches, GDumb~\cite{prabhu2020gdumb} proposed a degenerate solution to the problem of online learning ignoring the stream data and learning only on the memory samples.
While these efforts have significantly advanced the field of OCL, they are mostly evaluated on benchmarks that do not reflect real-deployment conditions. First, these benchmarks heavily rely on artificially constructed small datasets with sudden shifts in classes. Second, these benchmarks are incapable of measuring whether models can rapidly adapt to new data under a fast-changing distribution shift, which is one of the main problems in classical online learning literature~\cite{shalev2011online}. There have been efforts to address the second limitation by using new metrics that measure test accuracy more frequently during training~\cite{caccia2022new, koh2021online}. However, these metrics capture the adaptation to held-out test data rather than to the incoming future data. To remedy these limitations, there has been a new surge of benchmarks proposing datasets and evaluation protocols, which we discuss next.

\vspace{2pt}\noindent\textbf{Online Learning for Rapid Adaptation.}
Recent OCL benchmarks, \eg CLOC~\cite{cai2021online} and CLEAR~\cite{lin2021clear}, introduce data ordered by timestamps, forming a temporal stream of evolving visual concepts over a long span of time. They demonstrate that their data has a natural distribution shift over time, requiring rapid adaptation to newer data. Additionally, they mimic the traditional online learning setup by measuring the capacity for rapid adaptation with the evaluation being done on future data from the stream. Our work extends the efforts in this direction. We adopt the CLOC benchmark and propose a more realistic real-time evaluation that encourages efficient learning. 
It is worth mentioning that \cite{hayes2022online} has explored certain aspects of real-time evaluation in continual learning. However, they focus on efficiency for embedded device deployment, rather than rapid adaptation. Moreover, while their work investigates the computational cost associated with each training method, they only report the cost as an evaluation metric. In contrast, we constrain the training procedure by each method's computational cost.
\section{Methodology}
\label{sec:methodology}

\begin{figure*}
\centering
\includegraphics[width=0.94\linewidth]{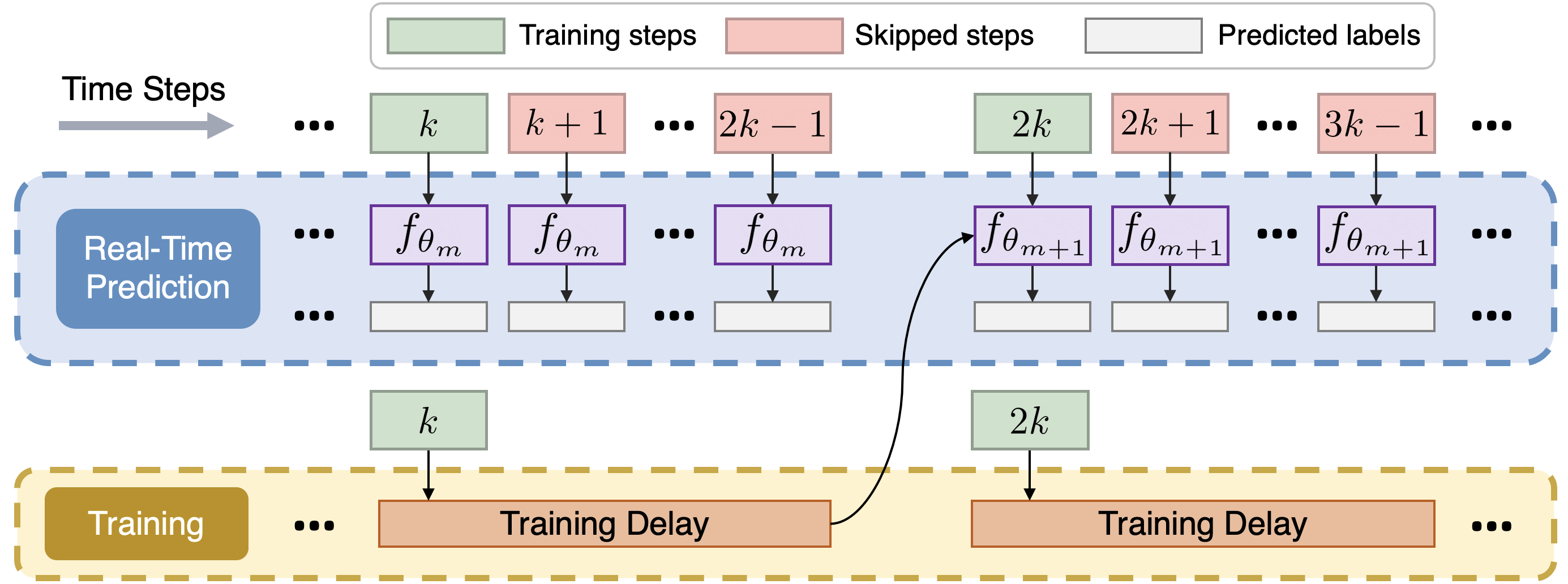}\par 
\caption{\textbf{OCL Real-Time Evaluation Setup.} In OCL, models perform real-time prediction on each revealed sample from the stream. At the same time, model training is performed on the revealed samples dictated by the method's delay. In this example, we demonstrate the evaluation procedure when the stream-model complexity $k$ is larger than 1. Due to the mismatch between the stream speed and the model computational cost, an "older version" of the model predicts samples while the model is being trained. Therefore, k-1 potential training batches are skipped for each training step.}
\label{fig:OCL_setup}\vspace{-0.3cm}
\end{figure*}

We start with the classical problem statement for online continual learning. Then, we formally introduce our proposed real-time evaluation that factors in training complexity through delayed evaluations. 

\subsection{Online Continual Learning}
\label{Sec_3_1_Traditional}

Online continual learning~\cite{shalev2011online,cai2021online} is the problem of learning a $\theta$ parameterized function $f_\theta : \mathcal{X} \rightarrow \mathcal{Y}$ that predicts a label $y \in \mathcal{Y}$ for an input image $x \in \mathcal{X}$. Unlike, classical supervised learning, the learning is performed on a distribution-varying stream $\mathcal{S}$ revealing data sequentially over steps $t \in \{1,2,\dots,\infty\}$. In particular, at every step $t$,  
\begin{enumerate}
    \item $\mathcal{S}$ reveals a set of images $\{x_i^t\}_{i=1}^{n_t} \sim \mathcal{D}_{j\leq t}$;
    \item $f_{\theta_t}$ generates predictions $\{\tilde{y}_i^t\}_{i=1}^{n_t}$ for $\{x_i^t\}_{i=1}^{n_t}$;
    \item $\mathcal{S}$ reveals true labels $\{y_i^t\}_{i=1}^t$;
    \item $f_{\theta_t}$ is evaluated by comparing $\{\tilde{y}_i^t\}_{i=1}^{n_t}$ to $\{y_i^t\}_{i=1}^{n_t}$;
    \item A learning method trains $f_{\theta_t}$, a criterion computes the training loss, then the parameters are updated to $\theta_{t+1}$.
\end{enumerate}
Note that $\mathcal{D}_{j\leq t}$ denotes a varying distribution that might not necessarily need to change at every stream step $t$. For example, if at step $t = 5$ we have $j=1$, this means the revealed data over all five previous steps is sampled from the same distribution $\mathcal{D}_1$. Moreover, observe that, unlike offline continual learning, online continual learning captures the capacity to adapt to new data since the prediction is performed before the training.

\vspace{2pt}\noindent\textbf{Key Issues.} As described earlier, the OCL evaluation in the literature~\cite{shalev2011online,cai2021online} overlooks the training complexity in step (5).
Under such a setup, all OCL methods have their parameters updated to $\theta_{t+1}$ before the stream $\mathcal{S}$ reveals images of the next step. Therefore, different OCL methods are evaluated irrespective of the training time and computational complexity at step (5). This is equivalent to evaluating different OCL methods on streams that present data at different rates.
In real-time settings, the stream reveals data at a rate that is independent of the training complexity of the OCL method. 
To that end, we propose a real-time evaluation paradigm for OCL that factors in training complexity through delayed evaluations.

\subsection{Real-Time Evaluation for OCL}\label{Sec_3_2_realtime}
As mentioned earlier, we need to define a notion of a fixed stream rate. That is to say, the stream reveals data at a fixed rate irrespective of how long OCL methods take to update the model in step (5). For streams that are twice as fast as it takes to train $f_\theta$, OCL methods will have their model updated on only 
half the number of revealed sets of images from the stream. Note that the latest updated version of $f_\theta$ will continue to provide predictions for newly revealed images during training even for those images the model did not train on. This setting reflects the realistic scenario of a server receiving  many queries uploaded by users, where each query needs a prediction irrespective of whether an in-house model is being trained. Given a stream $\mathcal{S}$ and an OCL method $\mathcal{A}$, we first define the notion of stream-model relative complexity $\mathcal{C}_\mathcal{S}(\mathcal{A}) \in \mathbb{R}^+$. In particular, for a stream-model relative complexity $\mathcal{C}_\mathcal{S}(\mathcal{A}) = 1$, the continual learning method $\mathcal{A}$ can update the model $\theta$ in step (5) before the stream $\mathcal{S}$ reveals the data of the next step. For any stream-model relative complexity $\mathcal{C}_\mathcal{A}(\mathcal{A}) = k > 1$, the stream is $k$-faster than the continual learning method $\mathcal{A}$. With this notation, we now propose our modified version of the OCL setting along with the corresponding real-time evaluation. Given a model $f_{\theta_m}$, a stream $\mathcal{S}$, and an OCL method $\mathcal{A}$ with $\mathcal{C}_\mathcal{S}(\mathcal{A}) = k$, at every step $t$,

\begin{enumerate}
\item $\mathcal{S}$ reveals a set of images $\{x_i^t\}_{i=1}^{n_t} \sim \mathcal{D}_{j \leq t}$;
\item $f_{\theta_m}$ generates predictions $\{\tilde{y}_i^t\}_{i=1}^{n_t}$ for $\{x_i^t\}_{i=1}^{n_t}$;
    \item $\mathcal{S}$ reveals true labels $\{y_i^t\}_{i=1}^t$;
    \item $f_{\theta_m}$ is evaluated by comparing $\{\tilde{y}_i^t\}_{i=1}^{n_t}$ to $\{y_i^t\}_{i=1}^{n_t}$;
    \item If $\text{mod}(t-1,k) = 0$, then the continual learner $\mathcal{A}$ completes updating $f_{\theta_m} \leftarrow f_{\theta_{m+1}}$ 
    and a new instance of training on $f_{\theta_m}$ commences (see Figure~\ref{fig:OCL_setup}).
\end{enumerate}

In this setting, OCL methods that are computationally more expensive to train will be updated fewer times. Therefore, for streams $\mathcal{S}$ where the distribution $\mathcal{D}_j$ changes rapidly (perhaps as often as every stream step), OCL methods with a large stream-model relative complexity may produce erroneous predictions, since the models are not updated enough and cannot adapt to the distribution change.

\vspace{2pt}\noindent\textbf{On the Computation of $\mathcal{C}_\mathcal{S}$.} Since $\mathcal{C}_\mathcal{S}$ only measures the relative complexity between the stream and an underlying continual learning method, we first introduce a minimal inexpensive OCL method as a baseline (denoted $\mathcal{A}$). For ease of comparison of the considered methods, we then assume that, due to the inexpensive nature of $\mathcal{A}$, online continual learning will be as fast as the stream where the stream-model relative complexity $\mathcal{C}_{\mathcal{S}} = 1$. In particular, we consider a baseline that mitigates forgetting by simply storing and replaying recently seen samples. Then, given any other OCL method $\mathcal{B}$, we use the relative training FLOPs between $\mathcal{B}$ and $\mathcal{A}$ to determine $\mathcal{C}_\mathcal{S}(\mathcal{B})$. 
For example, ACE~\cite{caccia2022new} only modifies the loss objective of the baseline, and thus it is equivalent to the baseline in computational complexity. On the other hand, PoLRS~\cite{cai2021online} maintains and performs operations on three copies of the model. These copies are trained on every incoming batch, making PoLRS require $3\times$ the FLOPs needed by the baseline. As noted earlier, a method of a stream-model relative complexity value of $3$ performs the update in step (5) once every three stream steps. This corresponds to a delay in model updates by two steps. In Table \ref{table:methods_delay}, we summarize the corresponding delay of several popular OCL methods in our real-time evaluation setup.

\vspace{2pt}\noindent\textbf{Fair Comparison of OCL Methods with Different $\mathcal{C}_\mathcal{S}$.}\label{sec:slow_stream} We proposed a realistic setup of evaluating continual learners based on their training complexity with respect to the stream speed. However, one might argue that more expensive training routines could be accelerated by deploying more computational hardware allowing them to train on each revealed sample from the stream. For example, while PoLRS~\cite{cai2021online} requires $3\times$ more FLOPs than the simple baseline, one could deploy $3\times$ more hardware to make $\mathcal C(\mathcal A)=1$ for PoLRS.
This setup mimics the scenario of having a slow stream that matches the speed of the more expensive training method.

While the aforementioned setup normalizes the computational requirement of a given learning method $\mathcal A$ to match the stream speed, one should allow simpler training methods the same computational complexity. To that end, we propose boosting simpler and faster training methods to match the computational requirements of more complex methods. For example, we boost the simple experience replay method by sampling a larger number of instances from the replay buffer at each step $t$ to match the FLOPs of other training schemes. We note here that although this modification is both simple and naive, we found empirically that it is sufficient to outperform \emph{all} considered OCL methods. We leave the rest of the implementation details to Section~\ref{sec:slow_stream_exps}.

\section{Experiments}
\label{sec:experiments}

We compare OCL methods under our proposed real-time evaluation setup given two stream speeds: fast and slow, which capture different application scenarios. 
We benchmark the majority of OCL methods, which we group into three categories. (1) \emph{Regularization-based} OCL methods regularize training by modifying the classification loss. For example, RWalk~\cite{chaudhry2018riemannian} uses a regularizer to penalize abrupt changes in the parameters of the model. ACE~\cite{caccia2022new} introduces an asymmetric loss to treat memory buffer samples differently from incoming stream samples, while LwF~\cite{li2017learning} distills knowledge from previous steps. (2) \emph{Learning rate scheduler} methods, in this case PoLRS~\cite{cai2021online}, dynamically adapt the learning rate to changes in the stream distribution. (3) \emph{Sampling-based} methods alter the strategy used to update the memory buffer, \eg GSS~\cite{aljundi2019gradient}, or change the memory retrieval strategy, \eg MIR~\cite{aljundi2019online}.

\vspace{2pt}\noindent\textbf{Datasets.} We use the large-scale online continual learning dataset CLOC~\cite{cai2021online}, which contains 39 million time-stamped images exhibiting natural distribution shifts. The task is to identify the geolocation of a given image where the total number of geolocation labels is 712. To ensure consistency with CLOC~\cite{cai2021online}, we adopt the same dataset split approach. Specifically, we use the first 5\% of the stream for hyperparameter selection, uniformly sample 1\% from the remaining stream to build a held-out set for measuring backward/forward transfer (reported in the appendix), and utilize the rest of the stream for training and online evaluation.
Similar to CLOC~\cite{cai2021online}, we compare OCL methods using the \emph{Average Online Accuracy} metric, which measures the ability of models to adapt to incoming stream samples\footnote{This metric should not be mistaken for a training accuracy, as it evaluates the model on the next \emph{unseen} training batch before the batch is used for model training.}.

\begin{table}
\caption{\textbf{Training complexity and delay of considered OCL methods.} $^{*}$Note that GSS has a complexity of 6.5 but we rounded it down to 6 to facilitate the setup. This rounding may give GSS a slight advantage, but nonethless, it is outperformed by the baseline, ER, in all experiments.}
    \centering
    \begin{tabular}{llcc} \toprule
      \textbf{CL Strategy} & \textbf{Method$(\mathcal A)$} & \textbf{$\mathcal{C}_\mathcal{S}(\mathcal A)$} & \textbf{Delay} \\ \midrule
      Experience Replay & ER~\cite{chaudhry2019continual} & 1 & 0 \\ \midrule
      \multirow{3}{*}{Regularizations} 
       & ACE~\cite{caccia2022new} & 1 & 0\\
       & LwF~\cite{li2017learning} & \nicefrac{4}{3} & \nicefrac{1}{3}\\
       & RWalk~\cite{chaudhry2018riemannian} & 2 & 1\\
      \midrule
      LR Scheduler & PoLRS~\cite{cai2021online} & 3 & 2\\  \midrule
      \multirow{2}{*}{Sampling Strategies} 
      & MIR~\cite{aljundi2019online} & \nicefrac{5}{2} & \nicefrac{3}{2}\\ 
      &  GSS~\cite{aljundi2019gradient} & \,\,6$^{*}$ & 5\\
      \midrule
      \bottomrule
    \end{tabular}\vspace{-0.2cm}
    \label{table:methods_delay}
\end{table}

\vspace{2pt}\noindent\textbf{Implementation Details.} 
At each step $t$ during the experiments, $\mathcal{S}$ reveals a set of $128$ images,
and OCL methods augment the training batch with another $128$ images sampled from a memory buffer. This routine is performed until a single pass over the stream is completed. We use ImageNet~\cite{deng2010imagenet} pre-trained  ResNet50~\cite{he2016resnet, cai2021online} as a backbone. We use SGD with a learning rate of $5 \times 10^{-3}$ for all OCL methods except for PoLRS~\cite{cai2021online}, which works best with a learning rate of $1 \times 10^{-3}$. 
Unless otherwise stated, we set the size of the replay buffer to $4\times 10^4$ for all considered methods.
We use the same hyperparameters reported in all corresponding papers with the exception of the regularization parameter for RWalk~\cite{chaudhry2018riemannian}. We find that their reported regularization $\lambda=0.1$ is too small for the CLOC dataset; we cross validate it and find that $\lambda=2$ works best. Our implementation extends the codebase of Avalanche~\cite{Lomonaco2021avalanche}, which is a popular continual learning framework. We leave all the remaining implementation details and ablations to the appendix.

\begin{figure}[t!]
\centering
\includegraphics[width=0.9\linewidth]{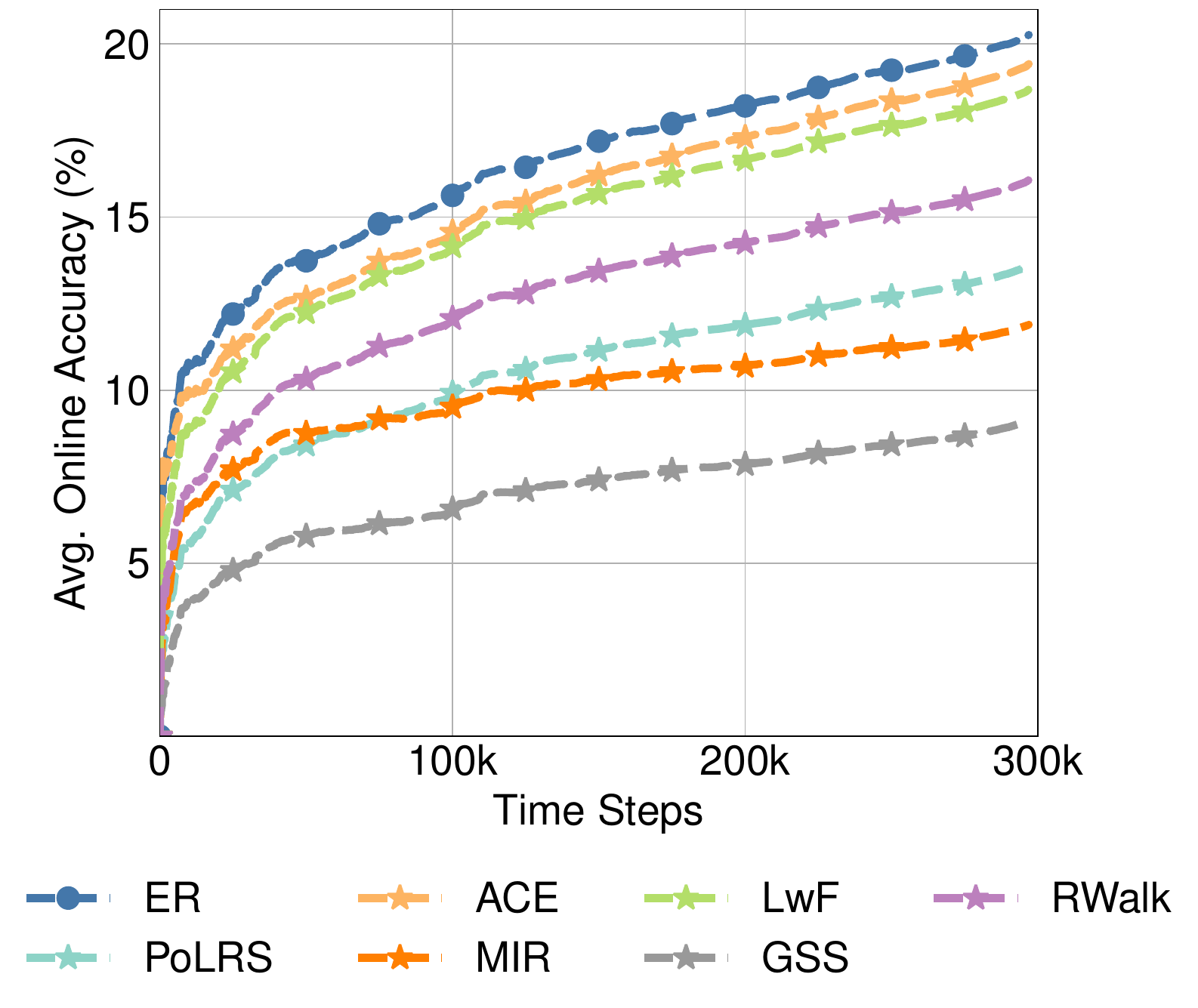}
\vspace{-0.2cm}
\caption{\textbf{Fast Stream Evaluation.} We compare the Average Online Accuracy of six methods from the OCL literature evaluated under the fast-stream setting. We observe that our inexpensive baseline, ER, outperforms all the considered methods. Surprisingly, the most computationally complex methods, MIR, PoLRS, and GSS, underperform the more efficient methods.} \vspace{-0.3cm}
\label{fig:fast}
\end{figure}

\subsection{Fast Stream: Real-Time Evaluation}\label{sec:fast_evaluation}
We first consider the fast-stream scenario, where the stream reveals data at a high rate. We consider the simple baseline \ER~\cite{chaudhry2019continual}, which performs one gradient step on a batch of $256$ ($128$ revealed by the stream and $128$ from memory) images at every stream step. The $128$ images from memory are sampled uniformly. For simplicity and ease of comparison, we assume that \ER has a stream-model relative complexity of $\mathcal{C}_{\mathcal{S}}(\text{\ER}) = 1$. 
Consequently, any OCL method that is computationally more expensive than \ER will not keep up with the stream speed, and thus will be forced to skip training on a fraction of the incoming stream samples. 

\begin{figure*}[t!]
\centering
\includegraphics[width=\textwidth,height=\textheight,keepaspectratio]{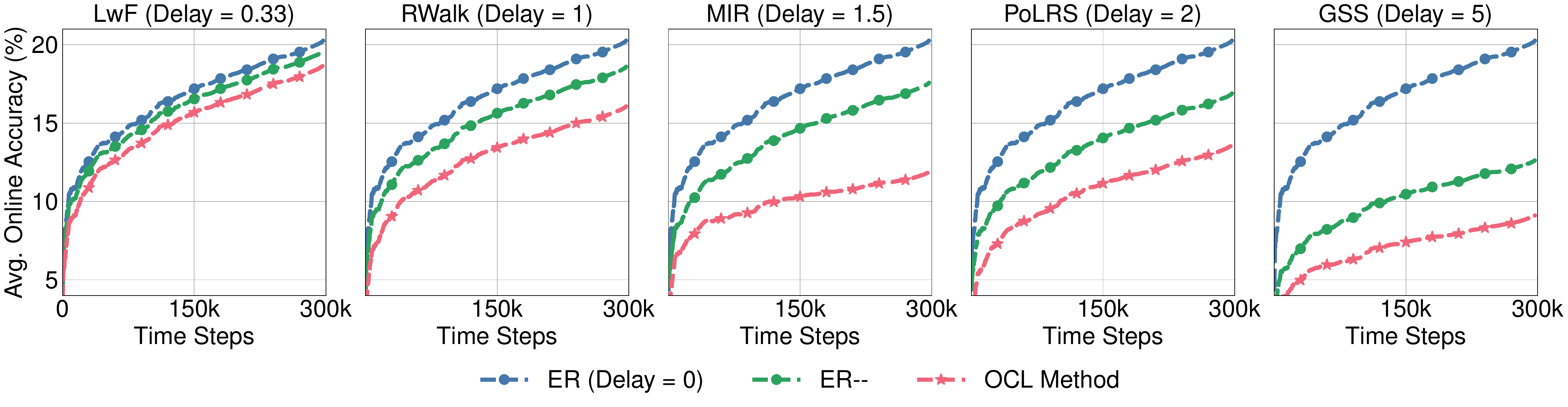} \vspace{-0.7cm}
\caption{\textbf{Fast Stream - Training Data Normalization.} We compare each method against \ER and its delayed version \ERMinusMinus. \ERMinusMinus performs extra gradient steps per time step to match the delay of the compared-against method, so it trains on fewer samples than \ER. We highlight that even when \ERMinusMinus is unfairly delayed to match each OCL method's delay, it outperforms all considered methods.}\vspace{-0.3cm}
\label{fig:fast-data-norm}
\end{figure*}
\vspace{2pt}\noindent\textbf{Benchmarking Relative Complexity of OCL Methods.} As discussed earlier, we use the total number of FLOPs required for each method to train on a stream as a proxy for the stream-model relative complexity $\mathcal{C}_\mathcal{S}$. We first compute the total FLOPs required to train \ER on a small subset of CLOC composed of $10^{3}$ images with a batch size of $10$. We then compute the corresponding FLOPs for various OCL methods and normalize them by the FLOPs of \ER. Since the backbone is shared across all methods, we further verify this normalized complexity by counting the effective number of forward and backward passes of each method. Because we assume that \ER is effectively as fast as the stream, methods that are twice more expensive in FLOPs have a complexity $\mathcal{C}_\mathcal{S}$ of $2$. We report in Table \ref{table:methods_delay} the stream-model relative complexity for OCL methods compared against \ER. Note that PoLRS is $3$ times as expensive as \ER since it requires training 3 different models to update the learning rate. 
In addition to the $128$ images revealed by the stream and the $128$ images sampled from memory at step $t$, GSS performs additional forward and backward passes on $10\times 128$ memory samples to ensure diversity when updating the memory buffer~\cite{aljundi2019gradient}. Therefore, GSS is roughly $6$ times more expensive than \ER. For each method in Table \ref{table:methods_delay}, we report the corresponding delay in stream steps, which is a consequence of some methods being slower than the stream. A delay of $2$ means that the stream will reveal $2$ steps worth of images before the model is updated.

\vspace{2pt}\noindent\textbf{Effect of Training Efficiency.} We start our analysis by investigating the effect of training efficiency on the performance of different OCL methods under the real-time evaluation setup. We plot the Average Online Accuracy curves per step in Figure \ref{fig:fast}, where the simple \ER baseline is in blue. Each method is evaluated according to its corresponding training delay reported in Table \ref{table:methods_delay}. Surprisingly, \ER outperforms all considered methods, and in some cases, by large margins. Interestingly, the most computationally intensive methods, MIR, PoLRS, and GSS, exhibit the lowest performance when real-time evaluation is considered. Specifically, the performance gap at the end of the stream reaches 11\%. In fact, we note that the Average Online Accuracy values of all considered methods, irrespective of how recently they were introduced to the literature, are approximately ordered based on their stream-model relative complexity $\mathcal{C}_\mathcal{S}$; the larger $\mathcal{C}_\mathcal{S}$ is, the worse the performance is under real-time evaluation. We attribute this trend to the fact that these inefficient methods may not be capable of keeping up with the fast-changing distribution $\mathcal{D}_j$ in the stream when they are subjected to a higher delay. For example, although PoLRS was proposed and tailored for the recent CLOC benchmark, it is significantly outperformed by the older method LwF when evaluated under the real-time paradigm. This hints that computational complexity can be a crutch for OCL methods learning on fast streams.
Our results show that while the current literature aims to improve learning by leveraging more expensive and advanced learning routines, the penalty incurred by delay overshadows any potential improvements gained by the algorithmic contributions. Notably, the state-of-the-art ACE method, which is as efficient as \ER and thus evaluated under no training delay, performs almost as well as the baseline. Additionally, ACE outperforms more expensive methods when evaluated on small-scale datasets (refer to Small-Scale Experiments in the appendix). Overall, our findings suggest that practical OCL methods, deployed in real-time applications, should prioritize efficient learning over expensive optimization.


\vspace{2pt}\noindent\textbf{Training Data Normalization.}
Previously, we have shown that delayed evaluation leads to  larger performance degradation as the delay gets longer. However, this could be attributed to the fact that OCL methods with larger delays end up effectively training on a fewer number of samples. This raises the questions: \emph{Do computationally more expensive methods with delays perform worse than \ER because they cannot cope with the changing distribution in the stream?}, or \emph{is the performance degradation due to expensive OCL methods being trained on effectively a fewer number of training examples?}
To address these questions, we conduct pairwise comparisons between \ER and each previously reported OCL method. Assuming that the stream speed is fixed, we modify \ER to match the computational expense of the respective method by training on the same amount of data for a longer duration. 
To achieve this computational matching, we subject \ER to a delay by performing additional gradient descent steps at each stream step.
We refer to this modified baseline as \ERMinusMinus. 
The number of additional updates in \ERMinusMinus is determined such that its delay matches the delay of the corresponding OCL method. This guarantees that \ERMinusMinus trains on the same number of training examples compared to other OCL methods.

Figure \ref{fig:fast-data-norm} shows comparisons of \ER and \ERMinusMinus to each of the considered OCL methods. Since the non-delayed \ER is more efficient than \ERMinusMinus, \ER consistently outperforms \ERMinusMinus, confirming that efficiency is key for the real-time evaluation of OCL. More interestingly, even though \ERMinusMinus matches the complexity of each compared-against method and is subject to the same delay, it still outperforms all considered OCL methods. Moreover, while the gap between \ERMinusMinus and expensive approaches, \eg GSS, is smaller than the gap to \ER, expensive methods still lag behind \ERMinusMinus by up to $3.5\%$. This demonstrates that the degraded performance of considered OCL methods is not due to the fewer number of observed training examples. On the contrary, expensive OCL methods seem to be unable to cope with distribution-changing streams. Under real-time evaluations of fast streams, simple methods such as \ER and \ERMinusMinus may be more suitable for real-world deployment.

\begin{figure*}[t!]
\centering
\includegraphics[width=\textwidth,height=\textheight,keepaspectratio]{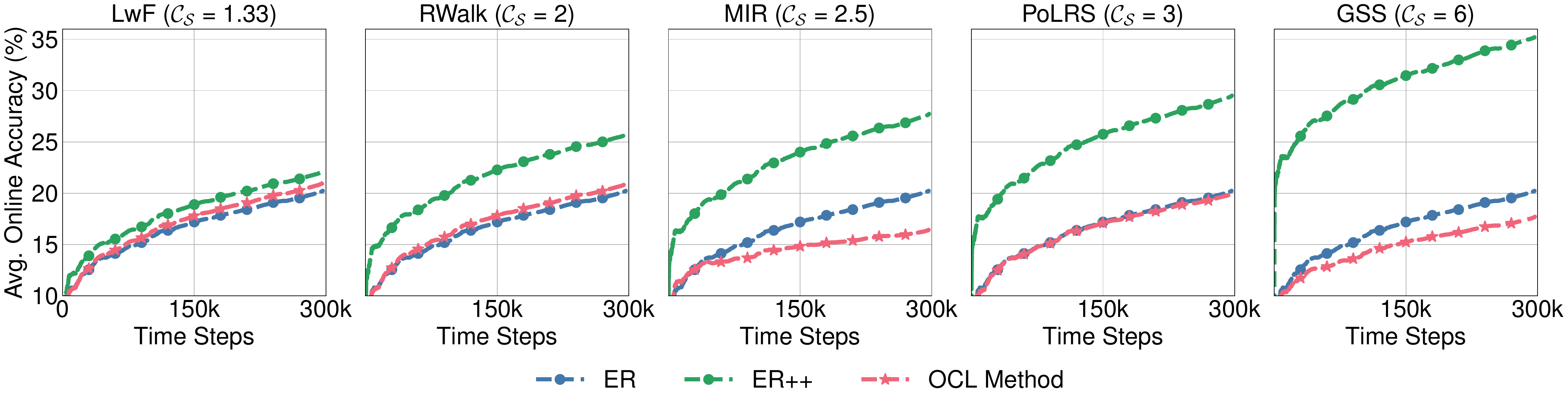} \vspace{-0.7cm}
\caption{\textbf{Slow Stream.} We compare each method against \ER and \ERPlusPlus, which performs extra gradient steps per time step to match the complexity $\mathcal{C}_{\mathcal{S}}$ of the compared-against method. \ERPlusPlus outperforms all the considered methods, sometimes by large margins.}\vspace{-0.2cm}
\label{fig:slow}
\end{figure*}

\subsection{Slow Stream: Complexity Normalization}\label{sec:slow_stream_exps}
In the fast stream setting, we considered the scenario where the stream is as fast as the \ER baseline, \ie, $\mathcal{C}_{\mathcal{S}}(\text{\ER}) = 1$. In this section, we consider streams that are as slow as the more expensive OCL methods. This setup motivates the following question: \emph{How do existing OCL methods perform if they were deployed on streams that match their training complexity?} 
Slower streams may enable expensive methods to train on the entire stream, without having to skip any stream steps. \emph{Would previously considered methods, which struggled in the fast-stream setting, be suitable for OCL deployment in this slow-stream scenario?} We compare various OCL methods \emph{under no delay} against their corresponding modified baseline \ERPlusPlus, which computes additional number of gradient steps per batch to match the complexity of the compared-against method, as determined from Table \ref{table:methods_delay}. This mimics the comparison on slow stream speeds, where no stream steps are skipped and comparisons are performed under \emph{normalized complexity}. Note that ACE only modifies the loss objective of \ER, so it matches the complexity of the baseline $\mathcal{C}_{\mathcal{S}}(ACE) = 1$. As a result, evaluating ACE in the slow-stream setting is identical to evaluating it in the fast-stream setting, which was already done in Figure~\ref{fig:fast}. Therefore, we do not compare to ACE again in the following experiments.

We report the comparisons in Figure \ref{fig:slow}, where \ERPlusPlus is shown in green and OCL methods in red. Moreover, we show \ER, which is unfavorably compared against both, in blue. 
While RWalk outperforms \ER in the slow stream setting, this is an unfair comparison, since  \ER is not utilizing the fact that the stream is slow and that it can benefit from further training. Once we simply add a few iterations to \ER, \ie, \ERPlusPlus, so as to match the training complexity of RWalk, \ERPlusPlus performs far better than RWalk. We find this to be consistent across all methods. 

Interestingly, computationally intensive methods, \eg, GSS and MIR, perform worse even when compared to \ER, which does not exploit the fact that the stream is slow. 
\ERPlusPlus is significantly better than GSS and MIR, by $17\%$ and $11\%$ respectively, expanding the performance gap to the baseline by around $15\%$ and $7\%$. These results question the suitability of the previously proposed methods to realistic OCL setups with large-scale datasets. Moreover, our results suggest that a computationally unfair evaluation setup (\ie not normalizing training complexity) in slow stream OCL can lead to misleading performance conclusions, as is the case with RWalk. 
We highlight that PoLRS was originally tested on a slightly different setup to the slow stream, which included extra information represented by a user album~\cite{cai2021online}. We find that when the album information is removed and their reported baseline, \ie, \ER, is tuned properly, PoLRS has a similar performance to \ER as shown in Figure \ref{fig:slow}. This is still surprising since PoLRS is three times more expensive than \ER. More surprisingly, when matching the training complexity of \ER to PoLRS on the slow stream, 
\ERPlusPlus outperforms PoLRS by $9.5\%$. These results on the slow-stream setting support our earlier observation that current OCL methods are not yet suited for practical deployment.

\subsection{Effect of Memory Size}
We conduct experiments to test whether existing methods can outperform \ER under different memory buffer sizes. We repeat the fast-stream experiment from Section~\ref{sec:fast_evaluation} with memory budgets of $1\times 10^4$, $2\times 10^4$, and $4\times 10^4$ samples. The results are summarized in Table~\ref{tab:memory}. A larger memory size leads to a higher Average Online Accuracy across all methods. 
However, we emphasize that the trend across methods aligns with our previous finding that expensive-training OCL methods perform worse than their less expensive counterparts.
This is evident from the ordering of the methods in Figure \ref{fig:fast}, which holds regardless of the memory budget. Notably, \ER outperforms all considered methods.

\begin{table}[t]
\caption{\textbf{Memory Budget Analysis.} We test the effect of varying the memory budget in the fast-stream setting. We observe that 1) increasing the memory size results in better performance regardless of the method, and 2) ER outperforms all the considered methods regardless of the memory budget.}\vspace{-0.1cm}
    \centering
    \begin{tabular}{@{}lccc@{}}
        \toprule
        \multirow{2}{*}{Method} & \multicolumn{3}{c}{Memory Size} \\ \cmidrule{2-4}
          & $1\times 10^4$ & $2\times 10^4$ & $4\times 10^4$ \\ \midrule
        ER~\cite{chaudhry2019continual}  & \textbf{18.09} & \textbf{19.37} & \textbf{20.27} \\ 
        \hdashline
        ACE~\cite{caccia2022new}    & 17.53 & 18.55 & 19.42 \\
        LwF~\cite{li2017learning}   & 16.87 & 17.87 & 18.69 \\
        RWalk~\cite{chaudhry2018riemannian} & 14.60 & 15.29 & 16.07 \\
        \hdashline
        PoLRS~\cite{cai2021online}  & 12.40 & 12.98 & 13.59 \\
        \hdashline
        MIR~\cite{aljundi2019online}    & 11.37 & 11.64 & 11.89 \\
        GSS~\cite{aljundi2019gradient}  & 8.01 & 8.59 & 9.10 \\
        \midrule
        \bottomrule
    \end{tabular}\vspace{-0.3cm}
    \label{tab:memory}
\end{table}

\section{Conclusion}
\label{sec:conclusion}

We proposed a real-time evaluation benchmark for online continual learning, including a fast stream that continues to reveal new data even when the model has not completed training. 
This realistic setup forces computationally intensive methods to skip more training samples, highlighting the differences in training complexity between the considered methods.
Our results show that, under this setting, all considered methods underperform the simple \ER baseline.
We also explored scenarios where we normalized the number of seen training data or the computational complexity per time step, leading to the same conclusion that current methods are not yet optimized for real-world deployment.

\vspace{2pt}\noindent\textbf{Acknowledgments.}
This work was supported by the King Abdullah University of Science and Technology (KAUST) Office of Sponsored Research (OSR) under Award No. OSR-CRG2021-4648, SDAIA-KAUST Center of Excellence in Data Science and Artificial Intelligence (SDAIA-KAUST AI), Saudi Aramco, and UKRI grant Turing AI Fellowship EP/W002981/1. We thank the Royal Academy of Engineering and FiveAI for their support. Ameya Prabhu is funded by Meta AI Grant No DFR05540.

{\small
\bibliographystyle{ieee_fullname}
\bibliography{egbib}
}

\clearpage
\appendix
\setcounter{figure}{0}
\setcounter{table}{0}
\section*{Appendix}
\label{sec:supplementary}

\section{Limitations and Future Work}
\label{sec:discussion}
In the main paper, we proposed a novel evaluation paradigm for real-time online continual learning, aimed at simulating real-world scenarios as closely as possible.
However, we recognize that there are few limitations that can be addressed in future work. Although \ER and \ERPlusPlus achieved the highest Average Online Accuracy among all other considered methods, their performance only reached up to 20\% and 35\%, respectively. This low performance suggests that there is room for improvement with better OCL methods. Furthermore, ER may have had an advantage in this setup due to potential temporal correlations in the CLOC dataset. One possible solution is to design new OCL datasets that have less temporal dependency. Additionally, while ER demonstrated the best online adaptation among other methods, it performed slightly worse in information retention (see backward transfer in Appendix D). Future work could explore OCL methods that can optimize both online adaptation and information retention.
\section{Pre-training Free Online Continual Learning}
\label{sec:scratch}

In real world-settings, continual learning proceeds a pre-trained model on locally annotated available data; as reported in all previous experiments in the main paper. However, we consider in this section a setting where models are continuously trained without any pre-training. We repeat our experiments on the CLOC dataset while training the ResNet50 backbone from scratch. We find that in this setting, a learning rate of $5 \times 10^{-2}$ works best for all methods except PoLRS, which works better with a learning rate of $10^{-2}$.

\begin{figure}[h]
    \centering
    \includegraphics[width=0.95\linewidth]{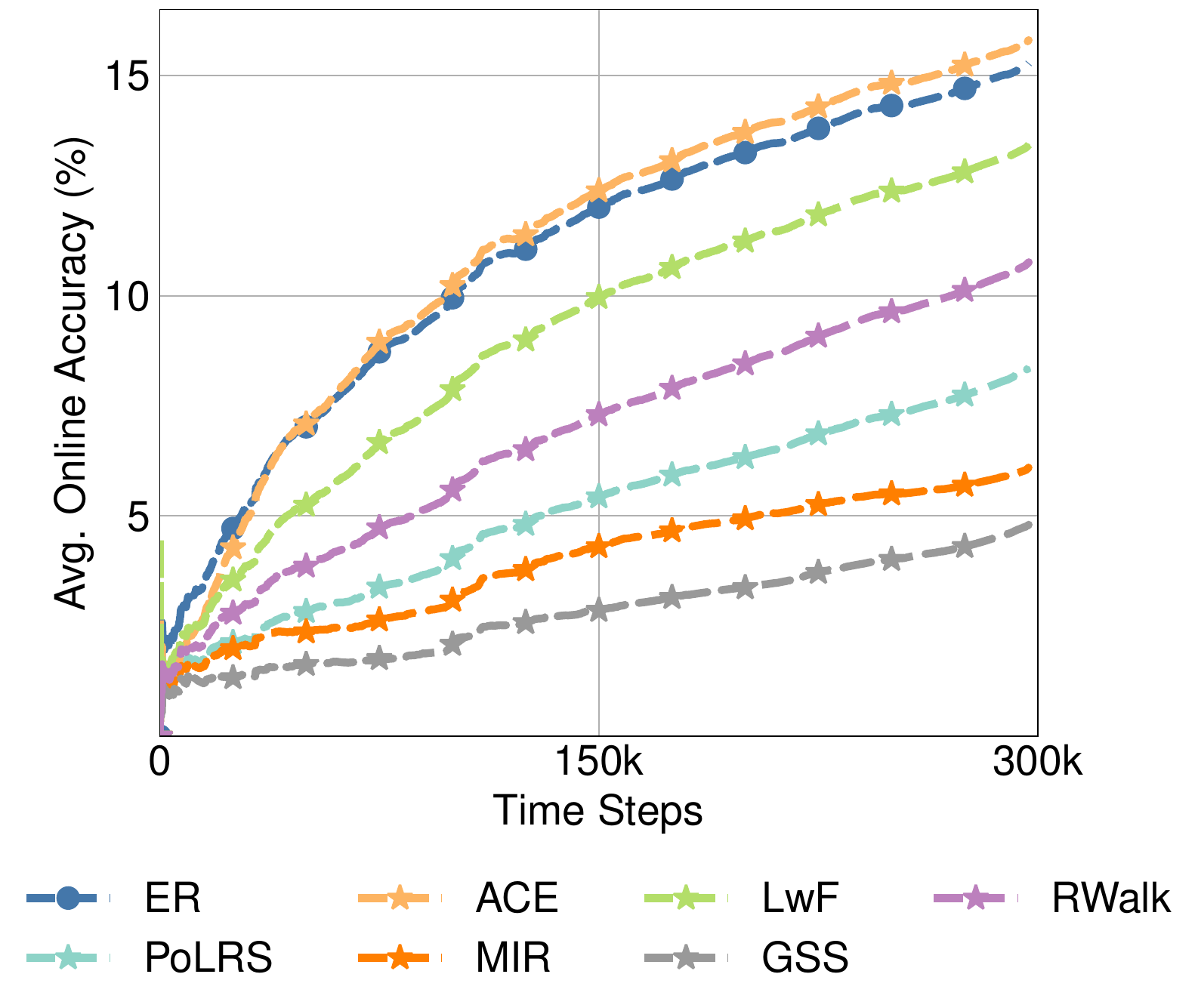}
    \caption{\textbf{Fast Stream Evaluation (Pre-training Free).} Consistent with the main paper experiments, the most efficient methods, \ER and ACE, significantly outperform the other considered methods, even when the backbone is trained from scratch.}
    \label{fig:fast_scratch}
\end{figure}

\begin{figure*}
\centering
\includegraphics[width=\textwidth,height=\textheight,keepaspectratio]{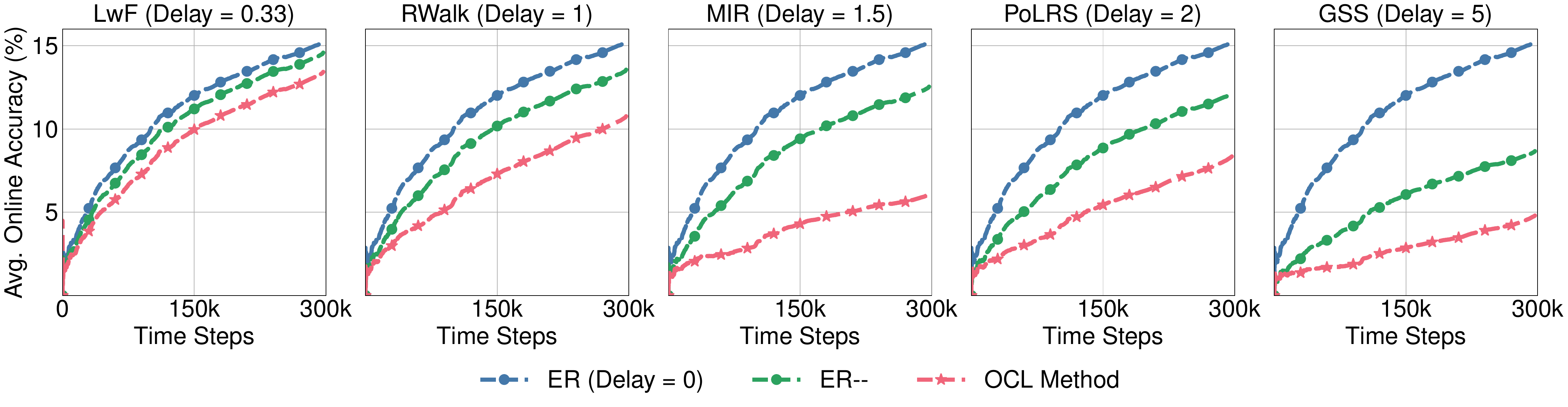} 
\vspace{-0.5cm}
\caption{\textbf{Fast Stream - Training Data Normalization (Pre-training Free)}. We compare each method against \ER and its delayed version \ERMinusMinus. \ERMinusMinus performs extra gradient steps per time step to match the delay of the compared-against method, so it trains on fewer samples than \ER. We highlight that even when no pre-training is deployed, \ERMinusMinus outperforms all considered methods.}
\label{fig:fast-data-norm-scratch}
\end{figure*}

\begin{figure*}
\centering
\includegraphics[width=\textwidth,height=\textheight,keepaspectratio]{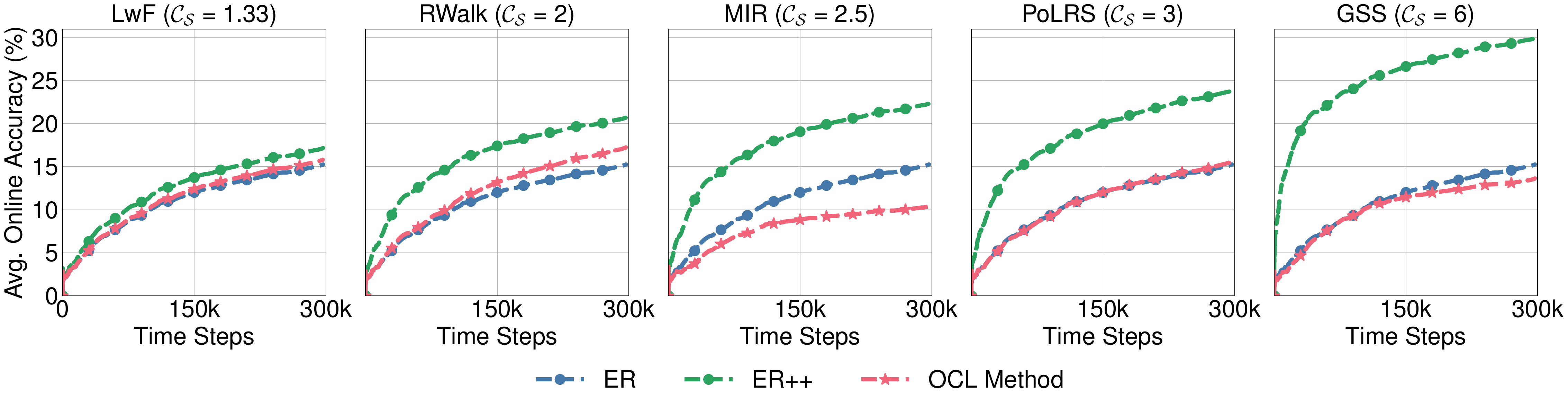}
\vspace{-0.5cm}
\caption{\textbf{Slow Stream Evaluation (Pre-training Free)}. We compare each method against \ER and \ERPlusPlus, which performs extra gradient steps per time step to match the complexity $\mathcal{C}_{\mathcal{S}}$ of the compared-against method. \ERPlusPlus outperforms all the considered methods, even when the backbone is not pre-trained.}
\label{fig:slow_scratch}
\end{figure*}

Following the evaluation grouping we use in the main paper, we present the results in three figures: Fast Stream (Figure \ref{fig:fast_scratch}), Fast Stream with Data Normalization (Figure \ref{fig:fast-data-norm-scratch}), and Slow Stream (Figure \ref{fig:slow_scratch}). We note that the results without pre-training reveal a similar conclusion to the main paper, where the simple baseline (\ER) outperforms more expensive OCL specific methods. Notably, ACE and \ER are as efficient as each other, and they both consistently outperform the more expensive methods. Interestingly, ACE only very marginally outperforms \ER despite that \ER does not use any specific OCL technique.

\section{Small-Scale Experiments}
\label{sec:small_datasets}

In this section, we test whether previous methods work well under real-time evaluation of continual learning if the dataset used is unrealistically small. To do this, we repeat the fast stream and slow stream experiments on CIFAR10 and CIFAR100 datasets, which many of the considered methods were originally tested on. We randomly shuffle these datasets without task boundaries constructing an online stream and sequentially train models over the stream in a single pass.

\vspace{2pt}\noindent\textbf{Implementation Details.} 
For all experiments on CIFAR10 and CIFAR100, the stream reveals $10$ images per time step which are augmented with another $10$ images sampled from a memory. This procedure creates a training minibatch of $20$ samples per time step. Moreover, we set the memory size to $100$. All models use a reduced ResNet18 as a backbone~\cite{caccia2022new} where we use SGD for optimizing models. For learning rate selection, we cross-validate on a held-out $5\%$ of each dataset. In CIFAR10, we find the following learning rate values to work best:
\begin{itemize}
    \item $10^{-3}$ for ER, ACE, LwF and RWalk
    \item $5 \times 10^{-3}$ for PoLRS, MIR, and GSS
\end{itemize}
As for CIFAR100, we use these learning rate values: 
\begin{itemize}
    \item $10^{-3}$ for ER, ACE, PoLRS
    \item $5 \times 10^{-3}$ for LwF, RWalk, GSS 
    \item $10^{-2}$ for MIR
\end{itemize}

\subsection{Fast Stream}
Recall that, for ease of comparison, we assume that \ER has a computational complexity that matches the stream speed and, thus, is able to train on all incoming samples in the fast stream. As a result, more computationally expensive methods than \ER lag behind the stream. This delay forces these methods to skip some incoming samples, where they effectively train on only a subset of the stream samples. In Table \ref{table:small_fast_stream}, we report the performance of considered OCL methods when evaluated on CIFAR10 and CIFAR100 datasets given a fast stream. We note that even when using small-scale datasets, ACE and \ER outperform all other considered methods in the practical, fast stream setting. Remarkably, ACE performance is only marginally better than \ER by less than 1.5\% on both datasets even though \ER does not do anything specific for continual learning.

\begin{table}[t]
\small
    \centering
    \begin{tabular}{@{}lcc@{}}
        \toprule
        \multirow{2}{*}{\textbf{Method}} & \multicolumn{2}{c}{\textbf{Average Online Accuracy (\%)}} \\ \cmidrule{2-3}
          & CIFAR10 & CIFAR100 \\ 
          \midrule
          ER & 51.60 & 13.62 \\ 
          \hdashline
          ACE & \textbf{53.05} & \textbf{14.42} \\
          LwF & 49.65 & 12.41  \\
          RWalk & 43.90 & 9.36 \\
          \hdashline
          PoLRS & 41.25 & 6.92 \\  
          \hdashline
          MIR & 43.55 & 10.57 \\ 
          GSS & 33.67 & 6.56 \\
          \midrule
        \bottomrule
    \end{tabular}
    \caption{\textbf{Small-Scale Experiments - Fast Stream.}  We repeat the fast stream experiments on the small-scale datasets CIFAR10 and CIFAR100. We notice that regardless of the size of the dataset, ACE and \ER outperforms all the considered methods.}
    \label{table:small_fast_stream}
\end{table}

\begin{table}[tb]
\small
    \centering
    \begin{tabular}{@{}llcc@{}}
        \toprule
         \multirow{2}{*}{$\mathcal{C}_{\mathcal{S}}$} & \multirow{2}{*}{\textbf{Method}} & \multicolumn{2}{c}{\textbf{Average Online Accuracy (\%)}} \\ \cmidrule{3-4}
          && CIFAR10 & CIFAR100 \\ 
          \midrule
          \multirow{2}{*}{\nicefrac{4}{3}}&LwF & 52.48 & 15.14  \\
          &ER++ & 54.09 & 14.79  \\
          \hdashline
          \multirow{2}{*}{2}&RWalk & 51.17 & 14.56 \\
          &ER++ & 56.71 & 16.35  \\
          \hdashline
          \multirow{2}{*}{\nicefrac{5}{2}}&MIR & 53.15 & 17.73 \\ 
          &ER++ & 58.03 & 17.37  \\
          \hdashline 
          \multirow{2}{*}{3}&PoLRS & 49.51 & 12.41 \\ 
          &ER++ & 59.52 & 18.45  \\
          \hdashline 
          \multirow{2}{*}{6}&GSS & 53.90 & 16.45 \\
          &ER++ & 59.76 & 17.09  \\
          \midrule
        \bottomrule
    \end{tabular}
    \caption{\textbf{Small-Scale Experiments - Slow Stream.} We repeat the slow stream experiments on the small-scale datasets CIFAR10 and CIFAR100. We note that in almost all cases, \ERPlusPlus outperforms the compared-against method. The only exceptions are LwF and MIR, and only when evaluated on CIFAR100. Even then, they have marginal gains over their corresponding simple baseline \ERPlusPlus, which indicates the need to develop better and more efficient methods for online continual learning.}
    \label{table:small_slow_stream}
\end{table}

\subsection{Slow Stream}
In this setting, similar to the main paper, the slow stream is as slow as the more expensive OCL method. That is to say, methods in this stream are able to train on all stream samples. Also recall that since methods have different training complexities in this stream, we use the modified baseline \ERPlusPlus for a fair comparison. \ERPlusPlus matches the complexity of each compared-against method by performing additional gradient updates.

We show the results of the slow stream evaluation in Table \ref{table:small_slow_stream} when considering small-scale datasets such as CIFAR10 and CIFAR100.
Similar to the slow stream results on the CLOC dataset, \ERPlusPlus outperforms all considered OCL methods when evaluated on CIFAR10.
As for CIFAR100, \ERPlusPlus also outperforms all considered methods except LwF and MIR, which are only slightly better than \ERPlusPlus by no more than 0.36\%.

\section{Performance on Held-out Samples}
\label{sec:measuring_forgetting}

\begin{figure*}
\centering
\includegraphics[width=\textwidth,height=\textheight,keepaspectratio]{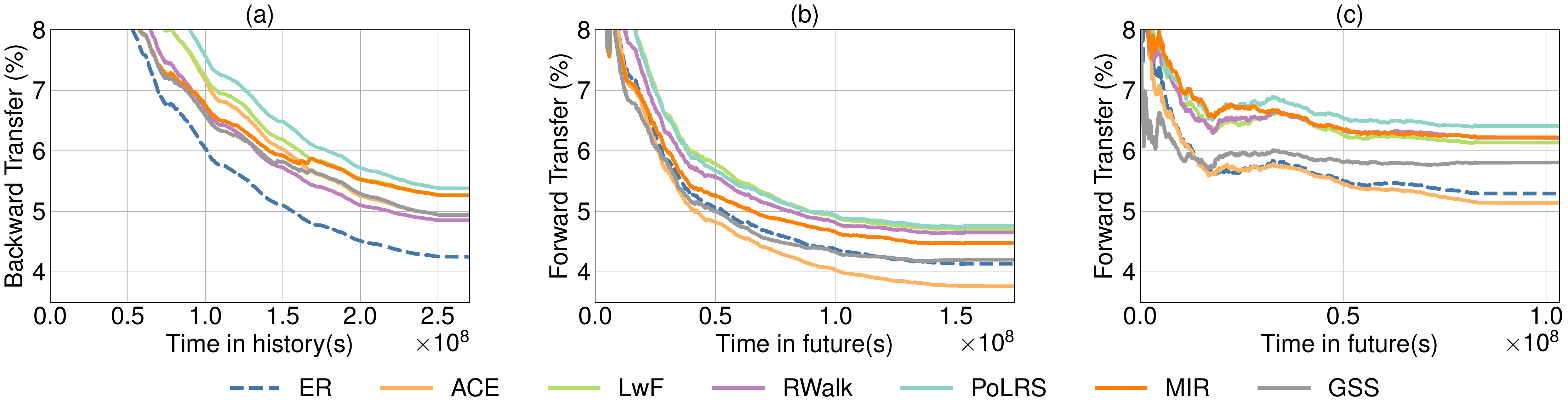}
\vspace{-0.5cm}
\caption{\textbf{Performance on Held-out Samples - Fast Stream}. (a) Backward transfer for a model obtained at the end of the stream. (b, c) Forward transfer for a model obtained at 1/3 and 2/3 of the stream respectively. We note that all methods have comparable performance to \ER in both backward and forward transfer metrics with only a maximum gap of 1.1\% across all plots. 
}
\label{fig:fast_other_metrics}
\end{figure*}

\begin{figure*}
\centering
\includegraphics[width=\textwidth,height=\textheight,keepaspectratio]{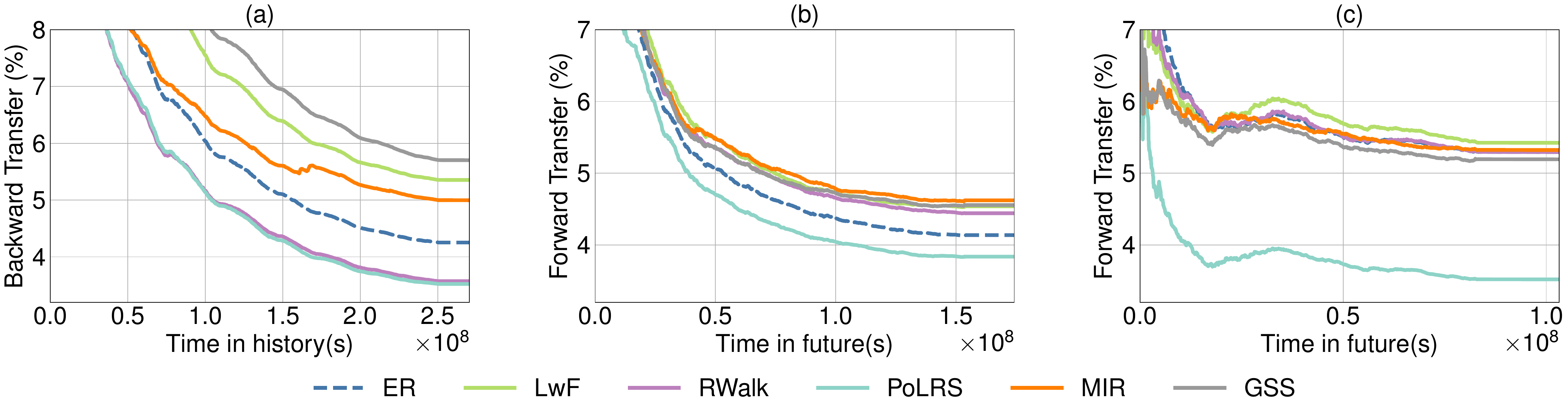}
\vspace{-0.5cm}
\caption{\textbf{Performance on Held-out Samples - Slow Stream}. (a) Backward transfer for a model obtained at the end of the stream. (b, c) Forward transfer for a model obtained at 1/3 and 2/3 of the stream respectively. We observe that all methods have comparable performance to \ER in both backward and forward transfer metrics. Even the most computationally intensive methods, namely GSS, and MIR, only outperform \ER in backward/forward transfer by a maximum of 1.5\%, which are the metrics they were optimized for. However, this minor advantage comes at the cost of much lower Average Online Accuracy. Note the ACE is not included in the plots because it has a training complexity of 1, and therefore, it was not evaluated in the slow stream.
}
\label{fig:slow_other_metrics}
\end{figure*}


In the main paper, we focused on the Average Online Accuracy as the main evaluation metric since it measures the adaptation to the next training batch  (\ie near future distribution), which is a highly desirable property in real-time streams with quick distribution changes. In this section, we consider two other evaluation metrics for continual learning, namely backward transfer and forward transfer. The backward transfer metric measures forgetting on past distributions, while the forward transfer evaluates models' adaptation to future distributions on held-out samples. It is important to note that forward transfer differs from Average Online Accuracy in that it evaluates the performance on held-out samples instead of future unseen training samples. Additionally, it monitors the performance on distributions that are far into the future, rather than the distribution of the next batch only.

To measure backward transfer and forward transfer, we uniformly sample and hold-out 1\% of the CLOC dataset from different time steps. The backward transfer of the final model obtained at the end of the stream steps is measured by computing the average accuracy on the unseen samples from past timesteps at every time step. Similarly, the forward transfer is evaluated by measuring the average accuracy on the unseen samples starting from the timestamp at which the model was obtained to the end of the stream. Following CLOC, we measure the forward transfer for the model saved at 1/3 and 2/3 of the stream.

Figures \ref{fig:fast_other_metrics} and \ref{fig:slow_other_metrics} show the backward and forward transfer results in the slow and fast streams, respectively. We observe that all OCL methods perform similarly to the \ER method, with only a slight advantage for some methods. For example, in the slow stream, although GSS and MIR require 6 and 2.5 times the computational cost of \ER, respectively, they only outperform \ER in backward transfer by 1.5\% and 0.7\%, respectively. Similarly, in the slow stream, they outperform \ER in forward transfer (at 1/3 of the stream) by only 0.4\% and 0.5\%, respectively. In the fast stream results shown in Figure \ref{fig:fast_other_metrics}, MIR and GSS also slightly outperformed \ER, even though they suffered from some training delay, indicating that they have a slight advantage in the forward/backward transfer metrics that they were optimized for. Importantly, this slight increase in GSS, and MIR performance comes at the cost of lower Average Online Accuracy, which is crucial in real-time continual learning. When considering Average Online Accuracy on both the slow stream and fast stream, the role of training efficiency becomes apparent.

It is worth noting that the backward transfer evaluation metric is slightly different from the objective of real-time online continual learning. In offline continual learning, methods are typically evaluated at the end of the stream on a held-out test set. However, this evaluation setup is irrelevant to real-time online continual learning, where streams constantly undergo rapid distribution shifts. Due to this stream dynamic, performance on samples from previous distributions is not a good indicator of whether models will make correct predictions on new incoming samples. This is because, in OCL with a naturally distributed stream, only the current stream samples are relevant for evaluation. For example, consider a naturally distributed stream revealing images of electronic devices. It is less likely for a stream to reveal images of 1995 devices as opposed to images of devices from 2022. This is the classical motivation behind online continual learning from a naturally distributed stream as opposed to offline continual learning. 

As for the forward transfer metric, we note that it may not be an accurate indicator of model's ability to predict future samples, especially in streams with rapidly changing distributions. Since forward transfer is calculated as a running average of accuracy over future distributions, it assigns equal weights to all future samples. However, this might be misleading in dynamic, real-time streams where near future distributions are more likely to be observed than those in the far future. In other words, a model that performs well on forward transfer may still struggle to adapt to the near future distribution. Therefore, while forward transfer can be a helpful metric in assessing a model's adaptability to future distributions, it should be interpreted with caution in real-time streams and supplemented with other evaluation metrics that focus on the near future distribution, such as the Average Online Accuracy.
\section{A Faster Stream}
\label{sec:faster_stream}

In the fast stream experiments reported in the main manuscript, we assumed that \ER is as fast as the stream for ease of comparison. As a result, \ER had a stream-model relative complexity of $\mathcal{C}_{\mathcal{S}}(\text{\ER}) = 1$, allowing it to train on the entire data stream. In contrast, methods with higher computational complexity than \ER had a stream-model relative complexity larger than 1, leading to some training delay. One could argue that our setup gave \ER an advantage by normalizing the stream speed to it. In this section, we explore a setting where the stream is faster than all evaluated methods.

\begin{table}[tb]
    \centering
    \begin{tabular}{llcc} \toprule
      \textbf{CL Strategy} & \textbf{Method$(\mathcal A)$} & \textbf{$\mathcal{C}_\mathcal{S}(\mathcal A)$} & \textbf{Delay} \\ \midrule
      Experience Replay & ER & 2 & 1 \\ \midrule
      \multirow{3}{*}{Regularization} 
       & ACE & 2 & 1\\
       & LwF & \nicefrac{8}{3} & \nicefrac{3}{2}$^{*}$\\
       & RWalk & 4 & 3\\
      \midrule
      LR Scheduler & PoLRS & 6 & 5\\  \midrule
      \multirow{2}{*}{Sampling Strategies} 
      & MIR & 5 & 4\\ 
      &  GSS & 12 & 11\\
      \midrule
      \bottomrule
    \end{tabular}
    \caption{\textbf{A Faster Stream Evaluation - Training complexity and delay of considered OCL methods when the stream speed is twice as fast as \ER.} $^{*}$Note that we rounded down the complexity of LwF from 1.67 to 1.5 for ease of implementation.}
    \label{table:methods_delay_2x}
    \vspace{-0.3cm}
\end{table}

\begin{figure}[tbh]
    \centering
    \includegraphics[width=0.95\linewidth]{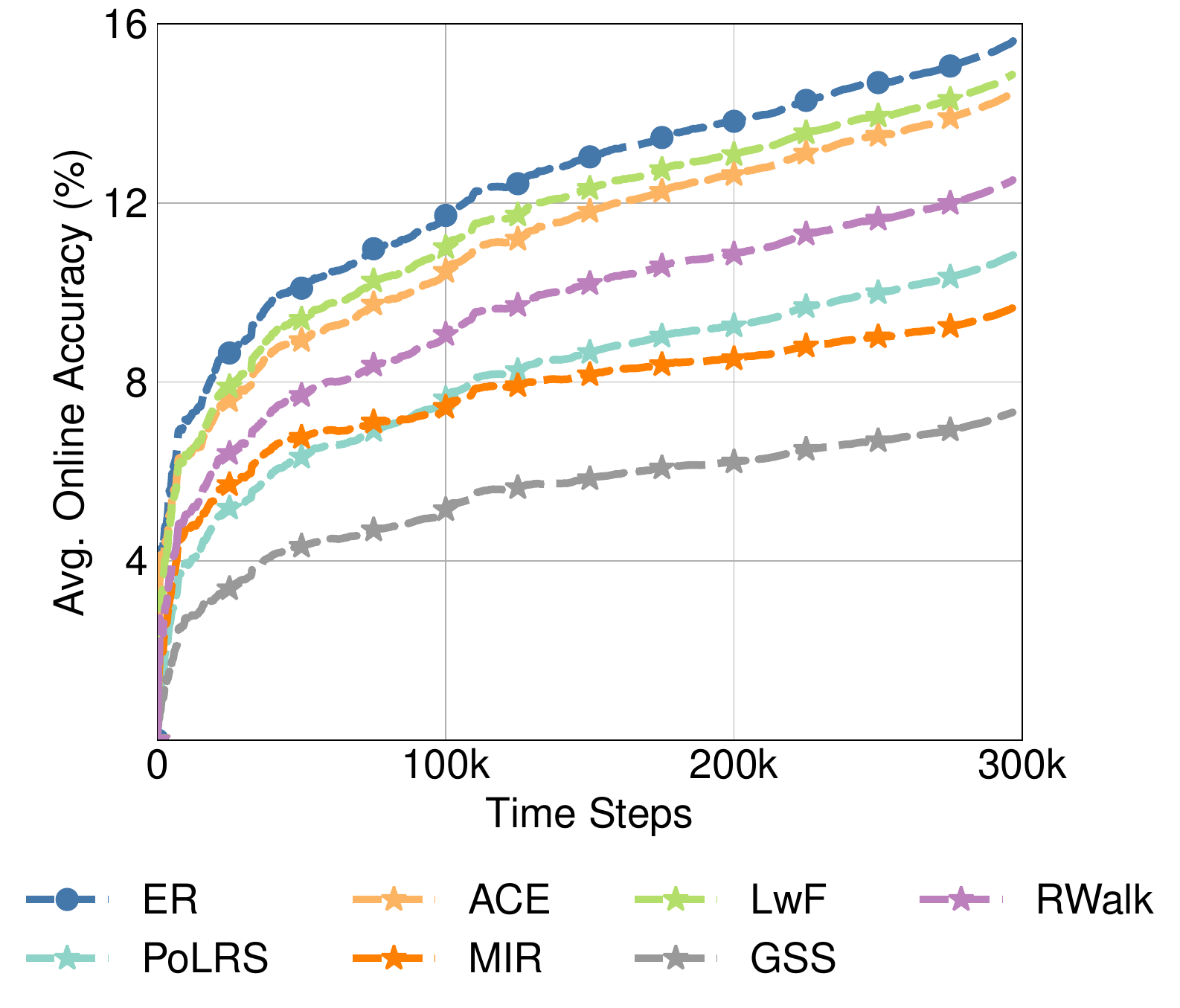}
    \caption{\textbf{A Faster Stream Evaluation.} We set the stream speed to twice as fast as \ER. Even when the stream speed is faster than \ER, \ER still outperforms all other considered methods, similar to the main manuscript experiments.}
    \label{fig:fast_2x}
\end{figure}

We repeat the fast stream experiments but with a stream speed that is twice as fast as \ER. Under this new stream speed, we outline in Table \ref{table:methods_delay_2x} the corresponding training complexity and delay for each considered method. In Figure \ref{fig:fast_2x}, we report the performance of considered OCL methods in the faster stream. Even when the stream is faster than \ER, we observe similar findings to the main manuscript. In particular, \ER still outperforms all other considered methods. 
\section{A Slower Stream}
\label{sec:slower_stream}

In the slow stream experiments presented in the main paper, we demonstrate that \ERPlusPlus outperforms all compared OCL methods. These findings raise the following question: \emph{What if the stream is slowed down to the point where more computationally expensive methods can perform additional gradient steps?} To explore this question, we allow each method to perform two Gradient Descent (GD) steps per time step. Accordingly, we adjust \ERPlusPlus to match the complexity of the boosted version of each method shown in Table \ref{table:methods_complexity_2x}. Note that performing two GD steps on LwF, RWalk, and PoLRS results in doubling their original complexity. In contrast, making an additional GD step on MIR and GSS only increases their complexity by one. This is because LwF, RWalk, and PoLRS require additional backward and/or forward passes with each GD step, while the sampling-based approaches, MIR and GSS, only require them with each new incoming data.

\begin{figure*}[t]
\centering
\includegraphics[width=\textwidth,height=\textheight,keepaspectratio]{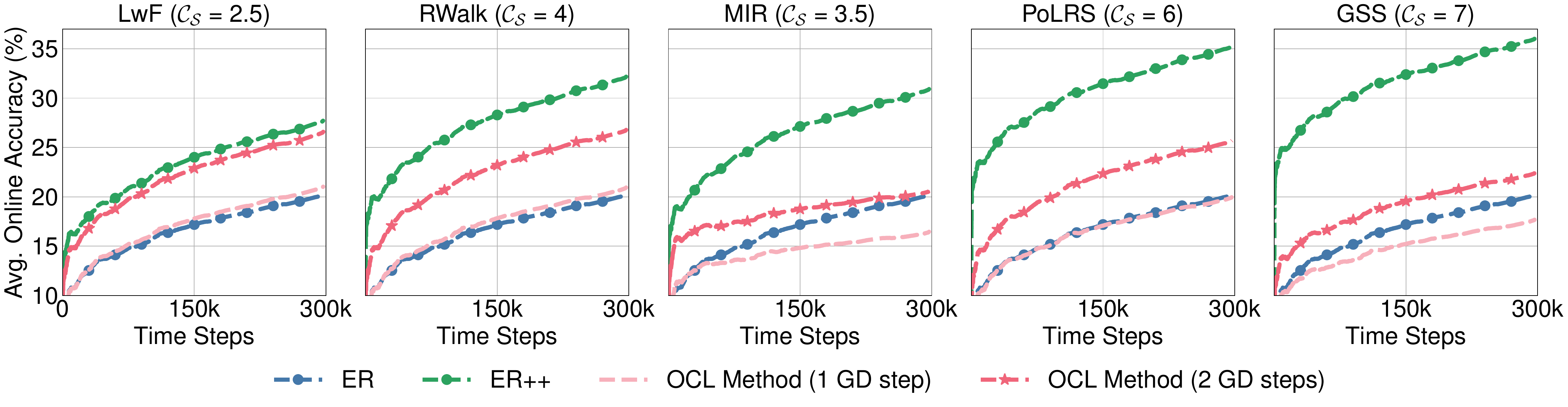}
\vspace{-0.5cm}
\caption{\textbf{A Slower Stream Evaluation}. We enable each OCL method to perform two GD steps per time step. We compare each method against \ER and \ERPlusPlus, which performs extra GD steps per time step to match the complexity $\mathcal{C}_{\mathcal{S}}$ of the compared-against method. For reference, we also show the performance of each OCL method with only a single GD step. Even when OCL methods perform additional GD steps, \ERPlusPlus still outperforms them all.}
\label{fig:slow_2x}
\end{figure*}

In Figure \ref{fig:slow_2x}, we report the performance of each OCL method with one and two GD steps per time step, where the results for one GD step are shown for reference only. We compare the boosted version of each method against \ER and its corresponding \ERPlusPlus. Consistent with the slow stream results presented in the main paper, \ERPlusPlus outperforms all the considered OCL methods. Compared to other methods, LwF has the smallest accuracy gap to \ERPlusPlus, suggesting that it is a highly efficient method. Although adding an extra GD step to GSS and MIR has reduced the previous gap to \ERPlusPlus from 17\% to 14\% and 11\% to 10\%, respectively, both methods lag behind LwF, which is a less computationally expensive method. In summary, the slower stream results indicate that \ERPlusPlus remains the optimal choice for achieving the highest Average Online Accuracy while using computational resources efficiently. 

\begin{table}[t]
    \centering
    \begin{tabular}{llcc} \toprule
      \textbf{CL Strategy} & \textbf{Method$(\mathcal A)$} & \textbf{$\mathcal{C}_\mathcal{S}(\mathcal A)$} \\ \midrule
      \multirow{2}{*}{Regularization} 
       & LwF & \nicefrac{5}{2}$^{*}$\\
       & RWalk & 4\\
      \midrule
      LR Scheduler & PoLRS & 6\\  \midrule
      \multirow{2}{*}{Sampling Strategies} 
      & MIR &  \nicefrac{7}{2}$^{*}$\\ 
      & GSS & 7$^{*}$ \\
      \midrule
      \bottomrule
    \end{tabular}
    \caption{\textbf{A Slower Stream Evaluation - Training complexity of considered OCL methods when each method is performing two GD steps per time step.} $^{*}$Note that complexities of these methods were rounded down for ease of implementation.}
    \label{table:methods_complexity_2x}
\end{table}

\section{Learning Rate Grid Search}
\label{sec:grid_search}

\begin{table}[tb]
    \centering
    \begin{tabular}{lcccc} \toprule
          LR & 0.001 & 0.005 & 0.01 & 0.05 \\ \midrule
          ER & 8.2 & \textbf{10.9} & 9.7 & 4.4 \\
          ACE & 8.1 & \textbf{8.8} & 7.9 & 3.4 \\
          LwF & 8.3 & \textbf{11.1} & 10.0 & 4.3 \\
          RWalk & 8.1 & \textbf{11.0} & 10.3 & 4.4 \\
          PoLRS & \textbf{10.8} & 10.6 & 9.3 & 3.9 \\ 
          MIR & 8.0 & \textbf{10.8} & 8.5 & 3.1 \\ 
          GSS & 7.5 & \textbf{9.8} & 9.0 & 3.6 \\
          \midrule
          \bottomrule
    \end{tabular}
    \caption{\textbf{Learning Rate Grid Search}. We perform cross-validation on each of the OCL methods considered, testing them on four different learning rate values. The Average Online Accuracy of the chosen learning rate value is denoted in bold for each method.}
    \label{table:lr_grid_search}
\end{table}

In order to determine the best learning rate for the methods in the slow and fast streams discussed in the main manuscript, we conduct a gird search across 4 values: \{0.001, 0.005, 0.01, 0.05\}, which we report in Table \ref{table:lr_grid_search}. We note that in practice, the stream speed is often unknown during training. Since the training delay is determined by the unknown stream speed, we conduct the grid search while assuming no training delay for all methods.

\end{document}